\documentclass{article}

\usepackage[nonatbib,preprint]{neurips_2019}

\usepackage[]{graphicx}
\usepackage{verbatim}
\usepackage{color}
\usepackage{cite}
\usepackage{subfig}
\usepackage[font={small}]{caption}
\usepackage[binary-units]{siunitx}
\usepackage{abbrevs}
\usepackage{pifont}
\usepackage{xspace} 
\usepackage{listings}
\usepackage[normalem]{ulem}
\usepackage{amsmath,amssymb,color,algorithm,algorithmic}

\definecolor{red3}{rgb}{0.80,0.00,0.00}

\makeatletter
\let\maybe@space@\xspace
\makeatother

\newcommand{\CM}[1]{}

\begin{document}

\date{}

\title{Breadth-first, Depth-next Training of Random Forests}

\author{
Andreea Anghel$^{*1}$\; Nikolas Ioannou$^{*1}${\color{white}\thanks{Equal contribution.}}\,\\\textbf{
Thomas Parnell$^1$\; Nikolaos Papandreou$^1$\; Celestine Mendler-D{\"u}nner$^2${\thanks{Work conducted while at IBM Research.}}\;\; Haris Pozidis$^1$}\\
 $^1$IBM Research, Zurich \\
  $^2$UC Berkeley, California\\
    \texttt{\{aan,nio,tpa,npo\}@zurich.ibm.com}\\  
    \texttt{mendler@berkeley.edu}, \texttt{hap@zurich.ibm.com}
}

\maketitle





\begin{abstract}
In this paper we analyze, evaluate, and improve the performance of training Random Forest (RF) models on modern CPU architectures.
An exact, state-of-the-art binary decision tree building algorithm is used as the basis of this study.
Firstly, we investigate the trade-offs between using different tree building algorithms, namely breadth-first-search (BFS) and depth-search-first (DFS).
We design a novel, dynamic, hybrid BFS-DFS algorithm and demonstrate that it performs better than both BFS and DFS, and is more robust in the presence of workloads with different characteristics.
Secondly, we identify CPU performance bottlenecks when generating trees using this approach, and propose optimizations to alleviate them.
The proposed hybrid tree building algorithm for RF is implemented in the Snap Machine Learning framework, and speeds up the training of RFs by $7.8\times$ on average when compared to state-of-the-art RF solvers (\texttt{sklearn}, \texttt{H2O}, and \texttt{xgboost}) on a range of datasets, RF configurations, and multi-core CPU architectures.

\end{abstract}

\section{Introduction}

Random forest (RF) models \cite{Breiman2001jml} are a powerful tool in machine learning (ML) that are being used in many applications such as bioinformatics \cite{boulesteix2012overview}, climate change modeling \cite{prasad2006newer} and credit card fraud detection \cite{bhattacharyya2011data}.
Their widespread usage stems from a number of advantageous properties.
RF models are amenable to high degree of parallelism, typically tend to have good generalization capabilities, natively support both numeric and categorical data, and allow interpretability of the results. A RF model is an ensemble model that uses decision trees as the base learners. 
Designing a scalable and fast tree-building algorithm is key for increasing the performance of RF models in terms of training time.
In particular, for large datasets and a large number of trees, to achieve good performance it is critical to design the tree-building algorithm with the characteristics of the underlying system in mind.

In this paper we evaluate the performance of different tree-building algorithms, namely breadth-first-search (BFS) and depth-search-first (DFS).
We investigate the trade-offs and identify the bottlenecks of both approaches.
Next, we propose a novel hybrid BFS-DFS algorithm, which can dynamically switch modes, and demonstrate that it performs better than both BFS and DFS, and further, it is more robust in the presence of workloads with different characteristics.
Moreover, we identify system-level bottlenecks at training time, and we alleviate them by (i) optimizing the layout of memory access pattern to be CPU cache friendly, (ii) employing explicit pre-fetching, and (iii) reducing the dynamic memory allocation overhead.
The proposed RF implementation scheme provides an improvement in training time of up-to $100\times$, and on average $7.8\times$ compared to state-of-the-art RF solvers (\texttt{sklearn}, \texttt{H2O} and \texttt{xgboost}), averaged over datasets, RF parameters, and two different hardware systems.


\section{Background}
\label{sec:background}

\textbf{Random Forest.} 
An RF model is an ensemble learning method which uses a decision tree as the base learner.
Training such a model consists of training a collection of N independent decision trees.
In a decision tree, each node represents a test on a feature, each branch the outcome of the test
and each leaf node (terminal node) a class label (for classification tasks) or a continuous value (for regression tasks).
For a standalone decision tree model, the tree is trained using all the examples and features present in the dataset.
Whereas, in the context of an RF model, each decision tree is trained using a bootstrap sample of the training examples.
In addition, when splitting each node in the tree, a different random subset of the features is used.
In order to identify the best split, one must identify the feature and feature value which, if split, will optimize a pre-defined metric.
For classification tasks, common choices for this metric include the Gini score as well as the binary entropy.
For regression tasks, it is common to use either the mean squared error or the mean absolute error.


\textbf{Data pre-sorting.}
Searching for the best split consumes the majority of the training time, and is thus the first part of the algorithm one should try to optimize.
A common solution is to pre-sort the training matrix for each feature \cite{mehta1996sliq, guillame2018arxiv, shafer96vldb}.
While this approach vastly reduces the complexity of finding the best split at each node, it introduces a one-off overhead: the time required to sort the matrix.
Whether this overhead can be amortized depends strongly on the tree depth as well as on the candidate features sampled at each split.
If the tree is grown to the point that all of the features have been sampled at least once, then sorting the matrix once in the beginning is more efficient than sorting at each node. 
To analyze this behaviour, we note that the question of how many features have been considered, for a given number of nodes in a tree, is equivalent to a variant of the well-known coupon collector's problem from probability theory~\cite{coupon14}.
An exact expression for the probability that all features have been used, and thus the cost of pre-sorting the matrix amortized, can be found in \cite{stadje90}.
To give an example: for a dataset with $30$ features, assuming $\sqrt{30}$ features are sampled at each node, it only takes a tree depth of $5$ before the probability that all features have been considered reaches $0.899$.
Moreover, if the pre-sorted matrix can be used across trees in a forest, its sorting cost is further amortized.
To this end, in our implementation we maintain a single read-only version of this sorted matrix in shared memory, used across all trees for the duration of the training.

\textbf{Different approaches to tree-building.}
The performance of training an RF model also strongly depends on the exact manner in which the tree is built, i.e., the order in which the nodes are traversed. 
One well-known approach is the so-called depth-first-search (DFS) algorithm \cite{cormen2009introduction}.
In DFS, after a node has been split, the tree-building algorithm recursively traverses the tree through the left-child.
Once a terminal node has been reached, it traverse up one level and recursively explores the right-child. 
A DFS-based RF implementation is offered by the widely-used machine learning framework, \texttt{sklearn}~\cite{pedregosa2011jmlr}.
An alternative approach to tree-building is to construct the tree \textit{level-by-level} using an algorithm known as breadth-first-search (BFS) \cite{mehta1996sliq}.
BFS is offered by software packages such as \texttt{xgboost} \cite{chen16kdd} and has recently been shown to work well when building trees on large datasets in a distributed setting\cite{guillame2018arxiv}. In what follows, we will compare BFS and DFS from a system-level perspective, in the context of training an RF model.

\section{Breadth-first, Depth-next tree building algorithm}
\label{sec:system}

In this section we will analyze the memory-access patterns of BFS and DFS, and propose a novel hybrid algorithm that is designed to deliver the \textit{best of both worlds}.


\textbf{Memory access patterns.}
\begin{figure}[t!]
  \centering
  \includegraphics[width=\columnwidth]{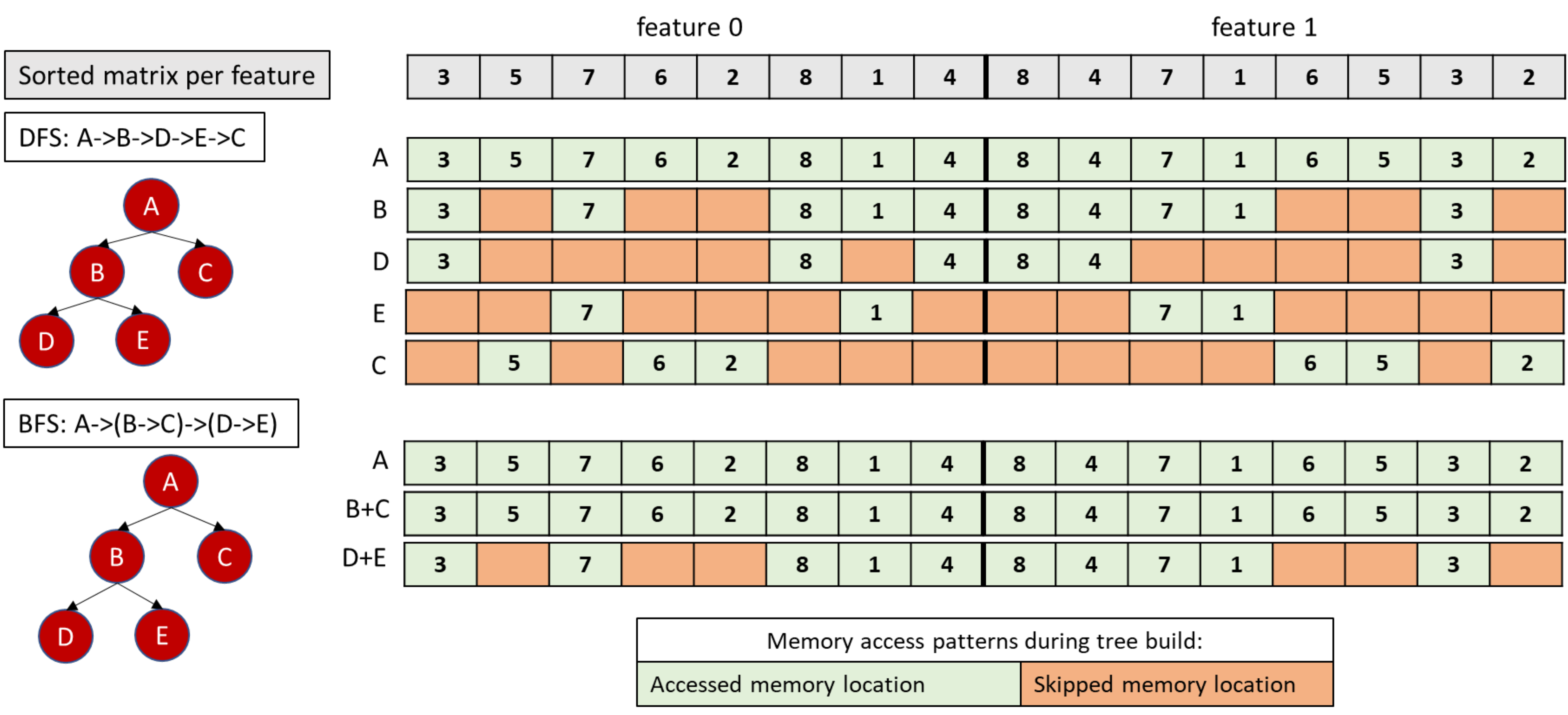}
  \caption{Comparing memory access patterns on the sorted data matrix between BFS and DFS on a dataset with 2 features and 8 training examples.}
  \label{fig:mem-acc-bfs-dfs}
\vspace{-0.2in}
\end{figure}
In the common case when the dataset does not fit in the CPU cache, accessing the data matrix in a cache-efficient manner is important to achieve high performance.
The notion of an \textit{active example} is critical to the analysis that follows.
We define an active example, at a given moment in the tree-building algorithm, to be any training example that is not associated with a terminal node.
A key insight is that at each tree depth level, most of the matrix (assuming most examples are still active) is accessed exactly once to compute the best split across the nodes of that depth.
A BFS tree-building algorithm, operating across all nodes at the same depth at each step, is thus well suited to access the data matrix in a cache-efficient manner.
DFS however is inherently less suited to exploit this property due to it repeatedly going down and up with respect to the tree depth as it builds the tree.

To illustrate the different memory access patterns of BFS and DFS, a tree with 5 nodes and a sorted matrix for a toy dataset with 9 examples and 2 features are shown in Fig~\ref{fig:mem-acc-bfs-dfs}.
Each item in the sorted matrix contains the example value and the example index (only the indices are shown for simplicity).
The expected memory access patterns for each step of the DFS and BFS algorithms are depicted below the sorted matrix; 
each row shows an algorithm step; green depicts an accessed memory location for this step; orange depicts a skipped memory location.
The example split illustrated results in leaf nodes C (examples 2,5,6), D (3,4,8) and E (1,7).
A DFS variant will start at node A, then proceed to nodes B, D, E, and C.
This process will result in a large number of skipped memory accesses, as shown in the figure.
In fact, DFS will quickly (w.r.t. tree depth) result in almost random accesses to the data matrix.
On the other hand, a BFS variant will start at node A, then proceed to nodes B, C, D, and E.
This, together with an additional optimization to compute all splits at each depth in one sequential access of the sorted matrix (see paragraph below on optimizations), results in a very cache-efficient memory access pattern to the matrix.

However, as the depth of the tree increases, and the number of active examples dramatically reduces, BFS no longer maintains a benefit over DFS: they both incur effectively random accesses to the sorted matrix, and exhibit very little re-use of cache lines brought to the CPU.
In fact, once there exist very few active examples, we expect the DFS to have better efficiency than BFS, especially if the active part of the sorted matrix (e.g., examples 1,3,7,8,4 at node B in Fig~\ref{fig:mem-acc-bfs-dfs}) fits in the CPU cache and is copied in a packed form to each tree node. DFS is guaranteed to only work with this active set of examples while expanding the tree from said node (e.g., starting at node B, discovering nodes D and E in Fig~\ref{fig:mem-acc-bfs-dfs}), thus exhibiting very good cache behavior.

\textbf{Breadth-first, Depth-next algorithm.}
Based on the above analysis, BFS is more cache-efficient at the first tree levels with most examples still active, while DFS would perform better towards the deepest end, when most examples are inactive.
Another argument for starting with a BFS approach is better cache re-use across trees in an RF, assuming trees are built in parallel: at low tree depths, where most examples are still active, each tree will read the sorted matrix sequentially from shared memory, and overlapping accesses across tree builders are very likely.
On the other hand, starting with a DFS approach would only have that benefit at the root node, after which each tree builder will quickly approach a random memory access pattern to the sorted matrix, resulting in dramatically reduced shared cache re-use across builders.
Due to the above, we have designed a Breadth-first, Depth-next tree building algorithm: we start with a BFS approach and at each BFS step we monitor the active number of examples; when the number of active examples is so small that we no longer expect BFS to be beneficial, we switch to a DFS approach; each node at the tree frontier proceeds with a DFS search for its own set of active examples.
The switching point is chosen when all the active data structures fit into the CPU cache size available to the each tree builder.
This hybrid algorithm is presented in Alg.~\ref{alg:hybrid-algo}, and has been implemented in the Snap Machine Learning framework~\cite{duenner18neurips} in C++.
To the best of our knowledge, this breadth-first, depth-next technique has not been applied in the context of RF before; the closest we could find was a hybrid BFS-DFS algorithm applied on the treewidth problem~\cite{zhou09socs}.
We implement multi-threading at the tree-level: each tree is trained in parallel on a different CPU core using OpenMP~\cite{dagum1998openmp} directives.
We also perform the sorting of the data matrix in a multi-threaded fashion during initialization.

\begin{algorithm}[t!]
    \begin{algorithmic}[1]
        \STATE sort training examples by feature \texttt{S[1:ft][1:ex]}
        \STATE \textbf{for each} tree \textbf{do}
        \STATE \quad randomly select a subset of training examples \texttt{E} with replacement
        \STATE \quad \textbf{while} (training stopping criteria not met) \textbf{do}
        \STATE \quad\quad execute one BFS iteration at current tree level \texttt{L} computing all splits across all nodes of \texttt{L}
        \STATE \quad\quad \textbf{if} (active data CPU cache size beneficial for DFS) \textbf{do}
        \STATE \quad\quad\quad \textbf{break}
        \STATE \quad \textbf{while} (training stopping criteria not met) \textbf{do}
        \STATE \quad\quad execute DFS for the remaining training examples at each node
    \end{algorithmic}
    \caption{Breadth-first, Depth-next Training Algorithm}
    \label{alg:hybrid-algo}
\end{algorithm}

\textbf{Further Optimizations.} 
During the BFS phase of the algorithm, we perform two main modifications: (i) the subset of features randomly selected are the same for each node at a particular depth (similar to ~\cite{chen16kdd}), and (ii) instead of building the tree in a node-to-example manner we do the opposite; at each tree level we sequentially walk the sorted matrix for all features chosen, maintaining an example-to-node mapping; by the end of this sequential scan, the splits for all nodes have been computed.
With the accesses to the sorted matrix being sequential, we have profiled the code and identified a lot of time spent accessing the example-to-node mapping, due to random accesses to it during the BFS.
We alleviated this performance issue by prefetching the subsequent example-to-node mappings (the indices of which are readily available in the subsequent entries of the sorted matrix).
The next performance issue that shows up in profiling is the memory accesses to the example label.
For binary classification problems, we exploit the fact that one bit is enough to hold the label information, and we pack that into the sorted matrix's example id field (using C bit fields), effectively stealing one bit from the id \textit{without} increasing the memory size of the matrix.
Applying all the above optimizations results in a performance profile dominated by vectorized floating-point instructions.

For the DFS phase of the algorithm, for each node we maintain a packed version of part of the sorted matrix corresponding to the node's active examples.
At each split, we copy the part of the parent's sorted matrix to the child that received the smaller number of examples after the split, then shrink the parent's sorted matrix and re-use it for the other child.
This optimization reduces the memory allocations (and de-allocations) needed at each DFS step by half compared to a straightforward implementation that allocates two new sub-matrices per split, copies the data over from the parent to the children, and then frees the parent's matrix.

\section{Evaluation}

In this section, we study the performance of our optimized RF implementation within the \texttt{Snap ML} framework in single-server environments.
For the remainder of the section we will refer to our implementation as \texttt{SnapRF}.

\textbf{Datasets.}
We will use three binary classification datasets to evaluate the performance of \texttt{SnapRF}:
Susy~\cite{Baldi2014SearchingFE} (5m examples, 18 features),
Higgs~\cite{Baldi2014SearchingFE} (11m examples, 28 features) and
Epsilon~\cite{epsilon} (400k examples, 2000 features).
In all cases, $75\%$ of the examples were used to construct a training set, and $25\%$ to form a test set.

\textbf{Infrastructure.}
For the following evaluation we used two multi-socket systems.
Firstly, a server with two 10-core Intel\textsuperscript{\textregistered} Xeon\textsuperscript{\textregistered} \mbox{E5-2630v4} CPUs, 125GiB RAM, running a 4.4 Linux kernel (Ubuntu 16.04).
Secondly, a server with two 20-core IBM POWER9 CPUs, 1TiB RAM, running a 4.15 Linux kernel (Ubuntu 16.04).
We disabled simultaneous multi-threading and fixed the CPU frequency to the maximum supported (2.2GHz for x86, and 3.8GHz for P9).

\textbf{Switching between BFS-DFS.}
Firstly, we evaluate the performance of \texttt{SnapRF} for BFS-only, DFS-only, and Breadth-first, Depth-next switching at fixed thresholds, and our automated cache-based switching mechanism. As explained in Sec.~\ref{sec:system}, for the latter, the switching from BFS to DFS occurs when all of the data structures corresponding to the active examples belonging to a node fit into the CPU cache size. The threshold is expressed as a percentage of the number of training examples.
If the fraction of active training examples in a given node is less than the specified threshold then the construction of the sub-tree originating from that node is performed using DFS.
The higher the threshold, the earlier the tree-building algorithm switches to DFS.

\begin{figure}[t!]
	\centering
	\includegraphics[width=\columnwidth]{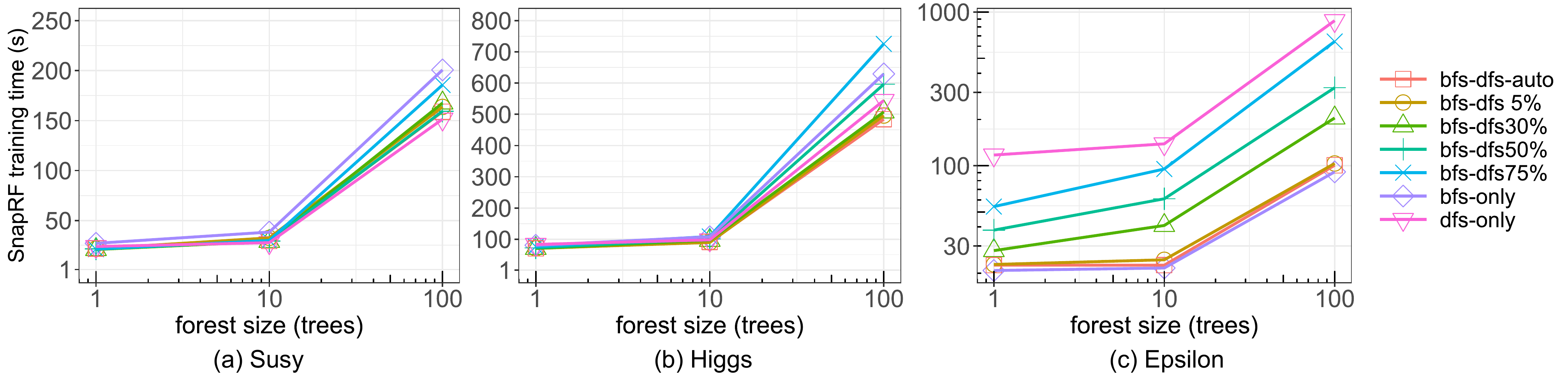}
	\caption{Performance of \texttt{SnapRF} for different BFS-DFS choices.}
	\label{bfs_dfs_th_analysis}
\end{figure}

In Figure~\ref{bfs_dfs_th_analysis} we show the training time as a function of the number of trees in the ensemble, assuming unbounded tree depth, for different BFS-DFS thresholds, and for the automated heuristic (\textit{bfs-dfs-auto}), on the x86 system.
We observe that for the Higgs dataset, BFS-only is the slowest choice, and the automated heuristic provides the best performance.
For Susy, DFS-only is the best choice, closely followed by our automated heuristic.
On the other hand, for the Epsilon dataset, we find that BFS-only actually performs the best, with the performance of the automated heuristic being again very close.
Compared to the other datasets, Epsilon has a very large number of features, and the BFS maintains a benefit to the optimization of using the same subset of features for all nodes at each depth (see Sec~\ref{sec:system}) overshadowing any DFS benefits.
Based on these results, we conclude that our automated switching heuristic is a robust choice that should provide good performance across a range of different datasets.
In all of the following results we will use this heuristic.

\textbf{Performance relative to baselines.}
We will now evaluate the performance of \texttt{SnapRF} relative to RF implementations offered in three widely-used ML frameworks: \texttt{sklearn}~\cite{pedregosa2011jmlr}, \texttt{H2O}~\cite{h2o} and \texttt{xgboost}~\cite{chen16kdd}\footnote{In all experiments we used \texttt{sklearn} version $0.19.1$, \texttt{H2O} version $3.26.0.3$ and \texttt{xgboost} version $0.9$}. We will evaluate the performance for (a) ensembles of 10 and 100 trees, (b) un-bounded trees and trees grown to a maximum depth of 20 and (c) on the x86 and P9 systems.

In all experiments and for all frameworks, we sample $\sqrt{m}$ features when splitting each node, where $m$ is the number of features in the dataset.
While \texttt{SnapRF} and \texttt{sklearn} both train each tree in the ensemble using a bootstrap sample of the training examples (i.e., sampling with replacement), 
\texttt{xgboost} and \texttt{H2O} train each tree by sampling without replacement from the set of training examples.
This difference is related to the fact that in the latter two frameworks, the underlying tree implementation is designed to also support \textit{boosting} \cite{friedman2001greedy}, for which sampling without replacement is common.
In order to compare the different frameworks as fairly as possible, we set the sub-sampling ratio in \texttt{xgboost} and \texttt{H2O} to $0.6321$: roughly corresponding to the probability of any given training example being included in a bootstrap sample when the dataset is large.
Note that since they use sub-sampling rather than bootstrap sampling, both \texttt{xgboost} and \texttt{H2O} train each tree using a smaller number of training examples, which should in principle allow them to run faster.
Since in this paper we are concerned with \textit{exact} tree-building (i.e., not using histogram-based techniques), we set the \texttt{tree\_method} parameter in \texttt{xgboost} to \texttt{exact}. Furthermore, we set the boosting-specific parameters \texttt{min\_child\_weight} and \texttt{lambda} parameters to zero, effectively disabling them.
In \texttt{H2O}, it is not possible to build trees using an exact method, therefore we can only compare with the histogram-based method that is used by default. 
Again, we note that by using histograms, one significantly reduces the complexity of searching for the optimal split and thus \texttt{H2O} should have a performance advantage in this regard.
In all training times reported, we do not include the time required to load the data from disk, nor do we count the time required to import the data into any framework-specific data structure (e.g. \texttt{H2OFrame} for \texttt{H2O} and \texttt{DMatrix} for \texttt{xgboost}).

\begin{figure}[t!]
	\centering
	\includegraphics[width=\columnwidth]{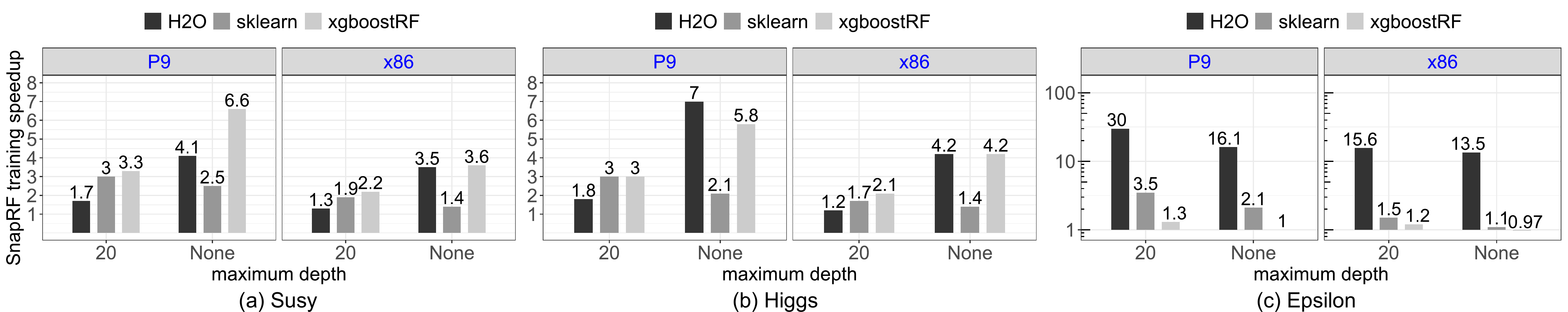}
	\caption{\texttt{SnapRF} training speedup over \texttt{sklearn}, \texttt{H2O} and \texttt{xgboost} for 10 trees on x86 and P9 systems.} 
	\label{benchmarking-10}
\end{figure}

In Figure~\ref{benchmarking-10}, we show the relative speed-up achieved by \texttt{SnapRF} over the other frameworks for 10 trees, for bounded and unbounded tree depths, and for the two systems under study. 
On the P9 system, averaging across datasets and tree depths, \texttt{SnapRF} shows an average speed-up of $2.7$x, $10.1$x and $3.5$x over \texttt{sklearn}, \texttt{H2O} and \texttt{xgboost} respectively. 
On the x86 system, \texttt{SnapRF} achieves an average speed-up of $1.5$x, $6.6$x and $2.4$x over \texttt{sklearn}, \texttt{H2O} and \texttt{xgboost} respectively. 

\begin{figure}[t!]
	\centering
	\includegraphics[width=\columnwidth]{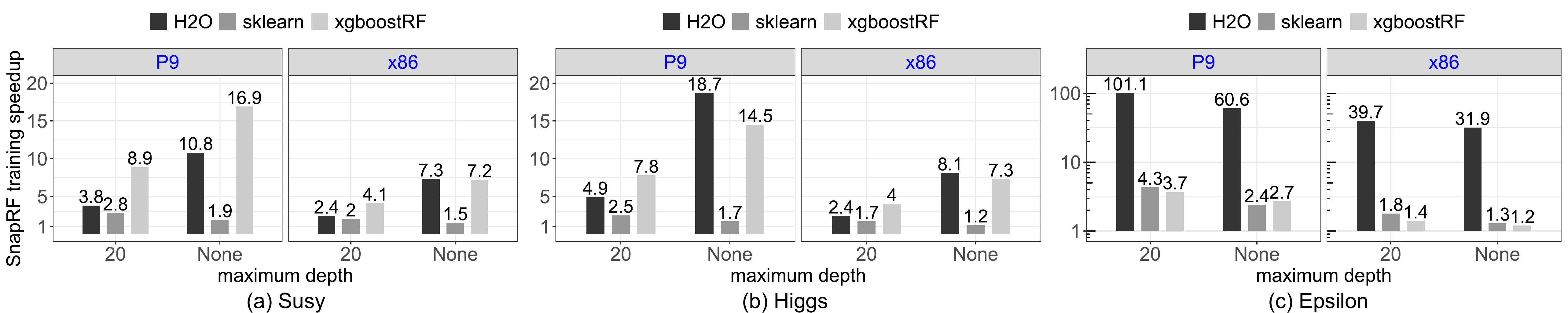}
	\caption{\texttt{SnapRF} training speedup over \texttt{sklearn}, \texttt{H2O} and \texttt{xgboost} for 100 trees on x86 and P9 systems.} 
	\label{benchmarking-100}
\end{figure}

To see how the implementations scale to larger ensembles, in Figure~\ref{benchmarking-100} we show the relative speed-up achieved when using 100 trees. 
The generalization accuracy achieved by all schemes is provided in Table~\ref{tbl:test-accuracies-100}.
On the P9 system, again averaging across datasets and tree depths, \texttt{SnapRF} shows an average speed-up of $2.6$x, $33.3$x and $9.1$x over \texttt{sklearn}, \texttt{H2O} and \texttt{xgboost} respectively.
On the x86 system, \texttt{SnapRF} achieves an average speed-up of $1.6$x, $15.2$x and $4.2$x over \texttt{sklearn}, \texttt{H2O} and \texttt{xgboost} respectively.
Thus, we find that the speed-up increases significantly when using a larger ensemble.

\begin{table}[]
	\centering
	\small
	\renewcommand{\arraystretch}{1.2}
	\begin{tabular}{|c|c|c|c|c|c|c|c|c|}
		\hline
		&  \multicolumn{4}{c|}{\texttt{max\_depth = 20}} &  \multicolumn{4}{c|}{\texttt{max\_depth = None}} \\ \hline
				\texttt{framework} & \texttt{snapRF} 	& \texttt{sklearn} 	& \texttt{H2O} & \texttt{xgbRF} & \texttt{snapRF} 	& \texttt{sklearn} 	& \texttt{H2O} & \texttt{xgbRF} \\ \hline \hline
				\texttt{susy} 			   & 0.801 & 0.801 & 0.801 & 0.801  &  0.8 & 0.799 & 0.799 & 0.8         \\ \hline
				\texttt{higgs} 			   & 0.737 & 0.737 & 0.736 & 0.738  &  0.751 & 0.75 & 0.751 & 0.751  \\ \hline
				\texttt{epsilon} 		 & 0.764 & 0.765 & 0.764 & \textcolor{black}{0.768}  & \textcolor{black}{ 0.758} & 0.759 & 0.761 & \textcolor{black}{0.766} \\ \hline
	\end{tabular}
	\captionsetup{justification=centering}
	\caption{Test accuracies across ML frameworks for 100 trees.}
	\label{tbl:test-accuracies-100}
\end{table}

\section{Conclusion}
We have designed a novel, hybrid BFS-DFS tree-building algorithm that is optimized for training RF models in modern multi-core CPU systems.
By dynamically exploiting different trade-offs at run-time, this hybrid algorithm is robust and able to outperform both BFS and DFS.
Moreover, we have performed a set of system-level optimizations that improve the memory access behavior of the algorithm and reduce its memory heap allocations.
The proposed hybrid BFS-DFS tree-building algorithm and optimizations have been implemented in the training routine of the RF model within the the Snap Machine Learning framework.
When compared against RF solvers from state-of-the-art ML frameworks (\texttt{sklearn}, \texttt{H2O}, and \texttt{xgboost}), across different CPU architectures and RF configurations, it provides a speedup in training time of up-to $100\times$ and on average of $7.8\times$.

{%
\footnotesize
\bibliographystyle{plain}
\bibliography{ms}
}

\end{document}